\documentclass[runningheads]{llncs}

\usepackage{amsmath,amssymb,mathtools,bm,etoolbox}

\DeclareMathOperator*{\argmin}{argmin}
\DeclareMathOperator*{\cel}{CrossEntropyLoss}
\DeclareMathOperator*{\glcml}{GLCMLoss}

\usepackage[T1]{fontenc}
\usepackage{graphicx}
\usepackage{array}
\usepackage{booktabs}
\usepackage{multirow}
\usepackage[normalem]{ulem}
\usepackage{adjustbox}
\usepackage[font=footnotesize]
{subfig} 
\usepackage[labelsep=period,font=small]{caption}
\usepackage{textcomp}
\usepackage{stfloats}
\usepackage{url}
\usepackage{verbatim}
\usepackage{cite}
\usepackage[table]{xcolor}
\usepackage{tabularx}
\usepackage{soul}
\usepackage{wrapfig}
\usepackage{orcidlink}
\usepackage{fancyhdr}
\newcommand{\fref}[1]{Fig.~\ref{#1}}
\newcommand{\tref}[1]{Table~\ref{#1}}

\let\llncssubparagraph\subparagraph
\let\subparagraph\paragraph
\usepackage[compact]{titlesec}
\let\subparagraph\llncssubparagraph
\titlespacing*{\section}{0pt}{1\baselineskip}{0.5\baselineskip}
\usepackage[margin=1.35in]{geometry}

\fancypagestyle{mystyle}{
    \chead{28$^{th}$ Pacific-Asia Conference on Knowledge Discovery and Data Mining (PAKDD 2024)}}

\begin{document}

\title{Adversarial-Robust Transfer Learning for Medical Imaging via Domain Assimilation}
%
%
\author{Xiaohui Chen\orcidlink{0009-0008-3521-0960} \and
Tie Luo\thanks{Corresponding author.}\orcidlink{0000-0003-2947-3111}}
%
\institute{CS Department, Missouri University of Science and Technology, Rolla MO 65409, USA
\email{\{xcqmk,tluo\}@mst.edu}} 

\maketitle
\thispagestyle{mystyle}

\begin{abstract}
In the field of Medical Imaging, extensive research has been dedicated to leveraging its potential in uncovering critical diagnostic features in patients. Artificial Intelligence (AI)-driven medical diagnosis relies on sophisticated machine learning and deep learning models to analyze, detect, and identify diseases from medical images. Despite the remarkable performance of these models, characterized by high accuracy, they grapple with trustworthiness issues. The introduction of a subtle perturbation to the original image empowers adversaries to manipulate the prediction output, redirecting it to other targeted or untargeted classes. Furthermore, the scarcity of publicly available medical images, constituting a bottleneck for reliable training, has led contemporary algorithms to depend on pretrained models grounded on a large set of natural images—a practice referred to as transfer learning. However, a significant {\em domain discrepancy} exists between natural and medical images, which causes AI models resulting from transfer learning to exhibit heightened {\em vulnerability} to adversarial attacks. This paper proposes a {\em domain assimilation} approach that introduces texture and color adaptation into transfer learning, followed by a texture preservation component to suppress undesired distortion. We systematically analyze the performance of transfer learning in the face of various adversarial attacks under different data modalities, with the overarching goal of fortifying the model's robustness and security in medical imaging tasks. The results demonstrate high effectiveness in reducing attack efficacy, contributing toward more trustworthy transfer learning in biomedical applications.

\end{abstract}
\section{Introduction}
Since its inception, Artificial Intelligence (AI) has evolved into a powerful tool across various domains. Particularly in the realm of medical diagnosis and treatment, AI has demonstrated impressive performance in predicting a range of diseases such as cancer, often treated as a classification problem. Beyond diagnosis, applying AI to medical problems for object detection and segmentation has also become focal points for researchers. The year 2018 marked a milestone toward real-world integration of AI in clinical diagnosis, where IDX-DR became the first FDA-approved AI algorithm designed for automatic screening of Diabetic Retinopathy.
On the research arena, numerous algorithms have been proposed that showcase remarkable performance in disease detection and diagnosis across diverse patient data modalities, such as MRI, CT, and Ultrasound. However, the lack of publicly available medical data, often attributed to privacy and the high cost of human expert annotation, remains a critical challenge for advancing medical AI research. In the meantime, model performance heavily depends on the size, quality, and diversity of training data to extract meaningful patterns. To address this challenge, researchers have turned to {\em transfer learning}, which takes models pretrained on large, extensive datasets of natural images and then fine-tunes the models' weights on the (smaller) medical datasets. 

However, natural and medical images have inherent differences from each other, forming a gap that is largely overlooked when applying transfer learning. This gap, which we refer to as ``domain discrepancy'', encompasses two particular aspects that eventually lead to heightened {\em vulnerability} of medical AI models to adversarial attacks. First, the {\em monotonic biological textures} in medical images tend to mislead deep neural networks to paying extra attention to (larger) areas {\em irrelevant} to diagnosis. Second, medical images typically have simpler features than natural images, yet applying overparameterized deep networks to learn such simple patterns can result in a sharp loss landscape, as empirically observed in \cite{Ma2021}. Both large attention regions and sharp losses render medical models trained via transfer learning susceptible to {\em adversarial examples}, the most common attack by adding small, imperceptible perturbations to original input images to induce significant changes in model output, resulting in mis-predictions. 

In this study, we introduce a novel approach called {\em Domain Assimilation} to bridge the gap between medical images and natural images. The aim is to align the characteristics of medical images more closely with those of natural images, thereby enhancing the robustness of models against adversarial attacks without compromising the accuracy of unaltered transfer learning. Our contributions are as follows:
\begin{itemize}
    \item We propose a novel domain assimilation approach embodied by a texture-color-adaption module integrated into transfer learning. This module transforms medical images to resemble natural images more closely, thus reducing the domain discrepancy.
    \item To prevent over-adaptation, which can lead to the loss of essential information in medical images and result in misdiagnoses, we introduce a novel Gray-Level Co-occurrence Matrix (GLCM) loss into the training process. This loss function incorporates texture preservation into the optimization process, ensuring the integrity of the original medical data, which is crucial for reliable diagnoses.
    \item We conduct extensive experiments across multiple modalities, including MRI, CT, X-ray, and Ultrasound, and evaluate the performance under various adversarial attacks, such as FGSM, BIM, PGD, and MIFGSM. Our results demonstrate that the proposed texture-color adaption with GLCM loss effectively enhances the robustness of transfer learning while maintaining competitive model accuracy.
\end{itemize}

\section{Related Work}
The inherent differences between medical and natural images pose challenges and some efforts were invested to narrow their disparities \cite{8302286, morra2020bridging}. A typical technique is grayscale image colorization, which is a prominent area in computer vision aiming to transform single-hue images into vibrant, colorful representations to enhance details and information for various applications. This field has garnered significant interest in the research community, finding applications in historical image restoration, architectural visualization, and the conversion of black-and-white movies into colorful ones, among others. Numerous existing works address colorization, either leveraging human-provided input such as scribbles or text, or utilizing color information from reference images \cite{7410412, huang2022}. Existing methods aim to predict the values of channels within chosen color spaces (e.g., RGB, YUV) by approaching colorization as a regression problem. In these colorization tasks, evaluation typically involves two steps: (1) convert color images into grayscale and generate colored output using designed algorithms; (2) use the original color images as ground truth for comparison with the generated output. However, tasks that need fine-grained colorization entail extensive data training, and one-on-one comparison to ground truth result in high computational overhead.

Furthermore, colorization of medical images poses additional challenges compared to natural images due to the absence of color space information in the original medical dataset. Prior research has underscored the importance of learning from natural images to address the domain gap with medical images. In a pioneering work \cite{8302286}, a bridging technique was employed, enabling the trained projection function from source natural images to transfer to bridge images derived from the same medical imaging modality. Subsequently, this transferred projection function was utilized to map target medical images to their respective feature space. In a subsequent study \cite{morra2020bridging}, a three-stage colorization-enhanced transfer learning pipeline was proposed. This involved allowing the colorization module to learn from a frozen pretrained backbone, followed by comprehensive training of the entire network on the provided dataset. The final step included training the network on its ultimate classification layer using a distinct dataset, aiming to enhance transferability. Recent work by Wang et al. \cite{Wang_Yan_2022} introduced a hybrid method incorporating both exemplar and automatic colorization for lung CT images. This approach utilized referenced natural images of meat, drawing inspiration from their similar color appearance to human lungs. Most recently, a self-supervised GAN-based colorization framework was proposed by \cite{app13053168}. This framework aims to colorize medical images in a semantic-aware manner and addresses the challenge of lacking paired data, a common requirement in supervised learning. 

While these existing approaches offer various advantages, many necessitate human intervention and rely on large sets of source or referenced images for learning color information to transfer to target medical images, resulting in a lengthy process. Additionally, texture information has been overlooked but is a crucial aspect in medical imaging tasks. 

\section{Preliminaries}
\begin{figure*}[!ht]
\centering
\subfloat[Brain MRI]{\includegraphics[width=0.5\textwidth]{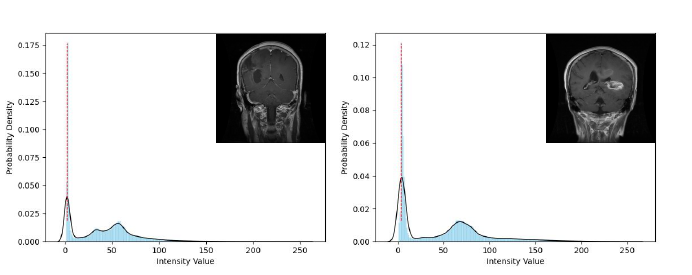}%
\label{fig1.0}}
\hfil
\subfloat[Breast Ultrasound]{\includegraphics[width=0.5\textwidth]{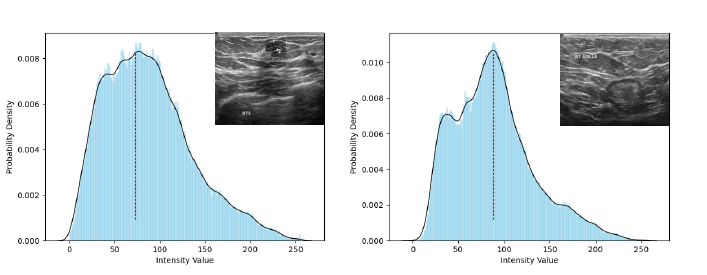}%
\label{fig1.1}}
\hfil
\subfloat[Chest CT]{\includegraphics[width=0.5\textwidth]{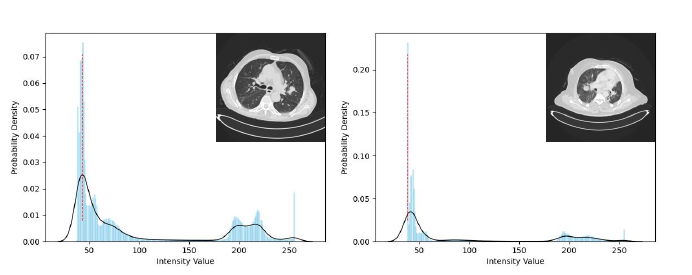}%
\label{fig1.2}}
\hfil
\subfloat[Chest X-Ray]{\includegraphics[width=0.5\textwidth]{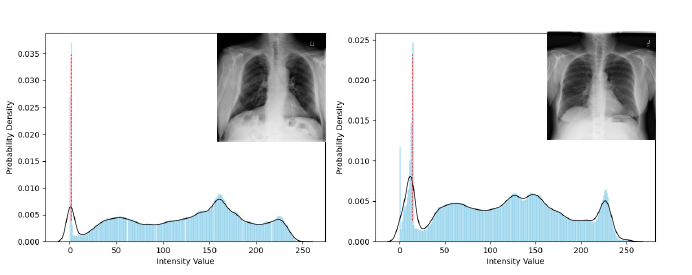}%
\label{fig1.3}}
\hfil
\subfloat[Tench, Tinca Tinca]{\includegraphics[width=0.5\textwidth]{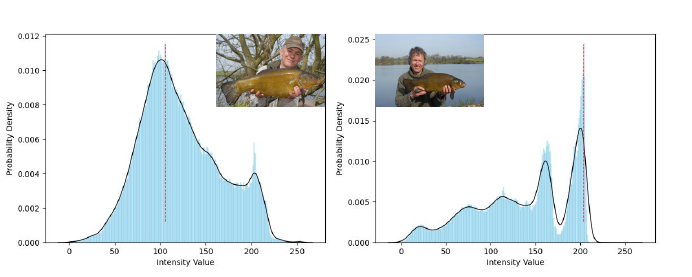}%
\label{fig1.4}}
\hfil
\subfloat[Goldfish, Carassius auratus]{\includegraphics[width=0.5\textwidth]{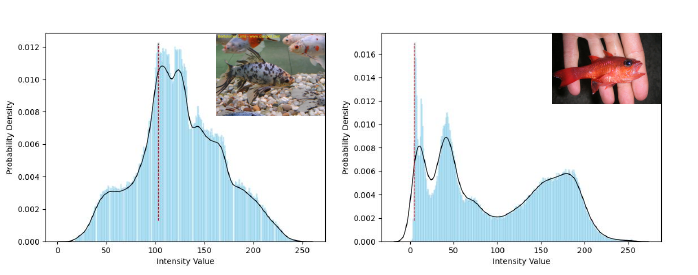}%
\label{fig1.5}}
\hfil
\subfloat[Goldfinch, Carduelis carduelis]{\includegraphics[width=0.5\textwidth]{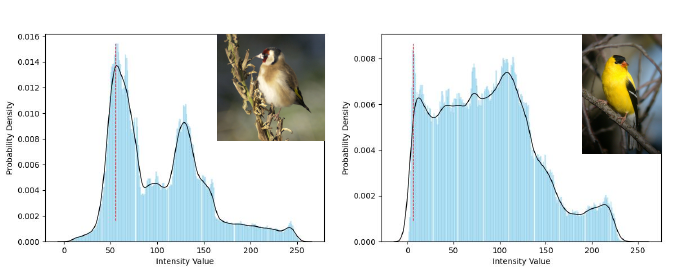}%
\label{fig1.6}}
\hfil
\subfloat[Japanese Spaniel]{\includegraphics[width=0.5\textwidth]{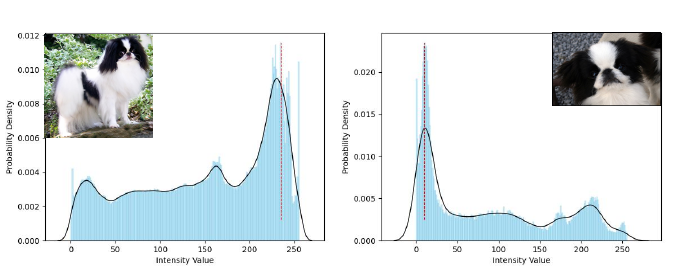}%
\label{fig1.7}}
\caption{Histograms showing pixel intensity value distribution of medical images and natural images. Conversion from RGB to grayscale was done for natural images for comparison purposes.}
\label{fig:hist}
\end{figure*}

\textbf{Medical Images vs. Natural Images}
Medical images typically manifest as grayscale, single-channel images, showcasing distinctive features from those found in natural images. The pixel-level values and spatial relationships within medical images convey inherent anatomical diversities among individual patients, often unveiling critical diagnostic features, for example, lesions. In contrast, natural images predominantly adopt the RGB color model, influenced by diverse illumination effects. Unlike the relatively standardized nature of medical images, natural images exhibit a broader spectrum of randomness and variability, encompassing a wide array of colors, objects, scenes, and textures. \fref{fig:hist} shows histogram comparisons between different imaging modalities \cite{Wang_2017,1jny-g144-23,Alyasriy2021-vw,ALDHABYANI2020104863} and natural images\cite{ILSVRC15}. A histogram is widely used in image analysis to understand pixel intensity value distribution. Here, we are plotting both frequency and probability density curve for a better demonstration. As we can see from \fref{fig:hist}a-d, within the same imaging modality, e.g., Brain MRI, the value distribution pattern is quite similar, however, the difference between medical images(a-d) and natural images(e-h) is rather apparent. We also compared natural images from different synsets in the well-known ImageNet2012\cite{ILSVRC15} dataset. As shown in the histograms, natural images intuitively contain more variations across different synsets and even within the same synset.

While first-order histogram-based statistics, such as mean, variance, skewness, and kurtosis, offer critical information on gray-level distribution, they lack the ability to depict spatial relationships at different intensity levels \cite{Agrawal_2012}. To further enhance the learning process, the importance of texture information within a medical image becomes apparent. Texture analysis involves scrutinizing the visual attributes, configuration, and distribution of elements constituting an object within an image. It delves into the spatial organization and recurrent patterning of pixel intensities, dispersed across the entirety of the image or specific regions. This interplay of intensities forms the fundamental essence defining the overall visual construct of the image \cite{varghese2019,Castellano2004}.

The Co-occurrence Matrix, also known as the Gray Level Co-occurrence Matrix (GLCM) \cite{4309314}, has been widely employed for texture analysis, especially on medical images \cite{9066263, singh2012classification, kumar2020feature}. The dimension of GLCM is defined by the number of intensity levels in an image; for example, a 4-value grayscale image would result in a $4 \times 4=16$ matrix. It serves as a texture descriptor, extracting second-order statistics and understanding the distribution of pairwise pixel graylevel values at a given distance and orientation, resulting in a specific offset. In the equation \eqref{eq1}, given an image I, $(i,j)$ represents the grayscale values, $(x,y)$ represents the spatial locations of pixels, and $(\nabla x, \nabla y)$ is the defined offset. In GLCM, common orientations include $0^{\circ}$(horizontal), $45^{\circ}$(front diagonal), $90^{\circ}$(vertical), $135^{\circ}$(back diagonal), and so on. 

\begin{equation}\label{eq1}
    C_{\nabla x, \nabla y}(i, j) = \sum_{x=1}^{n} \sum_{y=1}^{m} \begin{cases}
  1, & \text{if $I(x,y)=i$ and $I(x+\nabla x,y+\nabla y)=j$} \\
  0, & \text{otherwise}\end{cases}
\end{equation}
FROM GLCM we can further extract second-order statistics as follows. 

\textit{Angular Second Moment (ASM)} calculates the sum of the square of each value in the GLCM matrix and depicts the level of smoothness in an image.
\begin{equation}\label{eq:asm}
    ASM = \sum_{i,j}{P(i, j)}^2
\end{equation}

\textit{Contrast} quantifies how distinct the pixel intensity pairs are in terms of their differences. It reflects the amount of local intensity variation in the image.
\begin{equation}
    Contrast = \sum_{i,j}{(i - j)^2P(i, j)}
\end{equation}

\textit{Homogeneity} reflects the degree to which the pixel intensities in the image tend to be close to each other. A higher homogeneity value indicates that the pixel pairs in the image have similar intensity values, resulting in a more homogeneous and uniform appearance.
\begin{equation}
    Homogeneity = \sum_{i,j}\frac{P(i,j)}{1+|i-j|^2}
\end{equation}

\textit{Correlation} quantifies how correlated or linearly related the pixel intensities are in terms of their spatial arrangement. It indicates the degree to which the intensities at one pixel location can be predicted based on the intensities at another location.
\begin{equation}
    Correlation=\sum_{i,j}\frac{(i-\mu_i)(j-\mu_j)P(i,j)}{\sqrt{\sigma_i^2 \sigma_j^2}}
\end{equation}

\textit{Dissimilarity} reflects how dissimilar or different the pixel intensities are in terms of their spatial arrangement.
\begin{equation}\label{eq:dis}
    Dissimilarity=\sum_{i,j}P(i,j)|i-j|
\end{equation}

\textbf{Adversarial Attacks} In the domain of neural networks, particularly for a classification problem, we are presented with a set of input samples \(X \in \mathbb{R}^d\), each associated with ground truth labels \(y\) drawn from the distribution \(D\). Guided by the choice of the loss function \(L(\theta, x, y)\), where \(\theta\) denotes the model parameters subject to training, our objective is to identify the parameters that minimize the risk \(\mathbb{E}_{(x,y) \sim D}[L(\theta, x, y)]\). The challenge of adversarial robustness is cast as a saddle-point problem, as articulated in the work of Madry et al. \cite{madry2019deep} (see Equation \(\eqref{eq:1}\)), wherein the crux lies in solving both the inner maximization and outer minimization problem. In this study, we employ four gradient-based attacks as described in the following to assess the performance of our models.
\begin{equation}
\min_{\theta} \rho(\theta), \text{where } \rho(\theta)=\mathbb{E}_{(x,y) \sim D} [\max_{\delta \in S} L(\theta,x+\delta,y)] \label{eq:1}
\end{equation}

\textit{Fast Gradient Sign Method} (FGSM)
\cite{goodfellow2015explaining} introduced a single-step adversarial perturbation towards the input data with some magnitude along the input gradient direction. See \eqref{eq:2}.
\begin{equation}
X_{adv} = X + \epsilon * sign(\triangledown_{x} L(\theta, X, y)) \label{eq:2}
\end{equation}

\textit{Basic Iterative Method} (BIM) \cite{kurakin2017adversarial} proposed an iterative version of FGSM, also known as IFGSM, which perturbs the input data iteratively with smaller step size. See \eqref{eq:3}-\eqref{eq:4}.
\begin{align}
X_{adv}^0 &= X, \label{eq:3} \\ 
X_{adv}^{t+1} &= Clip_{x,\epsilon}{\biggl\{X_{adv}^{t} + \alpha * sign(\triangledown_{x} L(\theta,X_{adv}^{t},y))}\biggl\} \label{eq:4}
\end{align}

\textit{Momentum Iterative Fast Gradient Sign Method} (MIFGSM) \cite{dong2018boosting} proposed an extension of IFGSM that adds a momentum term to the optimization process to help attacks escape from local maxima. See \eqref{eq:5}-\eqref{eq:7}. 
\begin{align}
g_{0}&=0, X_{adv}^{0} = X,\ \alpha = \epsilon / T,\ T \text{ is the number of iterations} \label{eq:5} \\
g_{t+1} &= \mu \times g_{t} + \frac{\triangledown_{x} L(\theta, X_{adv}^{t}, y)}{\lVert \triangledown_{x} L(\theta, X_{adv}^{t}, y) \rVert_{1}},\ \mu \text{ is the decay factor} \label{eq:6} \\
X_{adv}^{t+1} &= X_{adv}^{t} + \alpha \times sign(g_{t+1}) \label{eq:7} 
\end{align}

\textit{Projected Gradient Descent} (PGD) \cite{madry2018towards} proposed an extension to fast gradient sign methods which projects adversarial examples back to the $\epsilon$-ball of x and is considered one of the strongest first-order attack. See \eqref{eq:8}-\eqref{eq:9}.
\begin{align}
X_{adv}^{0} &= X + \big(U^{d}(-\epsilon, \epsilon) \text{ if random start}\big), \label{eq:8} \\
X_{adv}^{t} &= \Pi_{\epsilon}\bigg(X_{adv}^{t-1} + \alpha \times sign(\triangledown_{x} L(\theta, X_{adv}^{t-1}, y))\bigg) \label{eq:9} 
\end{align}

\section{Method}
Medical images are monochromatic, single-channel grayscale representations. In the process of transforming them into RGB images, one can either generate values for distinct color channels or directly produce three-channel images by leveraging deep neural networks. Analogously to the colorization task, Convolutional Neural Networks (CNNs) can be employed to extract low-level information, such as textures, from images through operations like convolution and pooling.

\begin{figure*}[!ht]
\centering
\subfloat[Texture Module]{\includegraphics[width=0.5\textwidth]{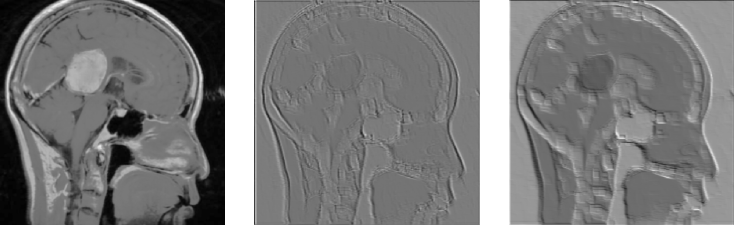}%
\label{fig2.0}}
\hfil
\subfloat[Color Module]{\includegraphics[width=0.34\textwidth]{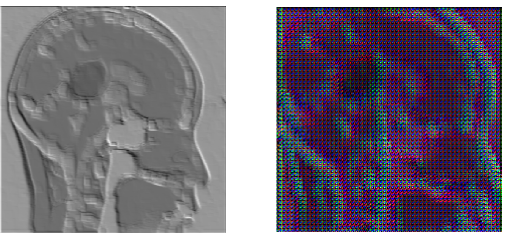}%
\label{fig2.1}}
\hfil
\caption{Example of the input and output of texture and color modules. (a) from left to right: original brain MRI image, encoder output, decoder output. (b) from left to right: input from texture module, three-channel image output.}
\label{fig:tex}
\end{figure*}

In the realm of transfer learning, we leverage a pretrained backbone obtained from an extensive dataset, utilizing it as a feature extractor to address specific tasks. The objective of this research is to facilitate the adaptive and voluntary learning of texture and color information by two distinct modules during the training process. This is achieved by employing a pretrained network and optimizing the three-channel output to enhance the classification performance. Inspired by previous work\cite{morra2020bridging}, we introduce lightweight texture and colorization modules preceding the pretrained backbone. The modules generate a three-channel image, subsequently fed into the pretrained models to produce a final classification outcome. See \fref{fig:tex} for a demonstration of the input-output-flow of the two modules.

\textbf{Problem Formulation} In the context of a generic classification problem, where the input $X \in R^d$, corresponding target labels $y$, and parameters $\theta$ are considered, our objective is to address the following challenge where our primary objective is to optimize the model parameters in order to minimize the loss between predicted and target labels.
\begin{equation}
    \theta = \argmin_{\theta} L(H(X, \theta), y)
\end{equation}

To effectively capture both texture and color information, our approach involves the integration of two distinct modules, dividing the overall network into four components: a texture module T, a color module C, a pretrained backbone B, and a final classifier F. Before the colorization process, it is imperative to preserve crucial texture features inherent in the input image. This is achieved by initially generating a single-channel image through the texture module, which is subsequently input into the color module. The color module's role is to learn three-channel information and output a colored image. Finally, the output is fed into the pretrained model for classification. This formulation of the problem leads to the following.
\begin{equation}
    {<\theta_T, \theta_C, \theta_B, \theta_F>} = \argmin_{\theta_T, \theta_C, \theta_B, \theta_F} L\Biggl(F\biggl(B\Bigl(C\bigl(T(X, \theta_T), \theta_C\bigl), \theta_B\Bigl), \theta_F\biggl), y\Biggl)
\end{equation}

Observing the proven efficacy of texture as significant features in medical image tasks such as classification and segmentation \cite{Castellano2004, Sharma_Ray_Sharma_Shukla_Pradhan_Aggarwal_2008, kiani2022texture}, it is noted that colorization, while enhancing the 'natural' appearance of medical images, may potentially distort the original texture information due to its fine-grained labeling at the pixel level. Despite its ability to improve alignment with pretrained models, we seek to mitigate the impact of colorization during the training process. To achieve this, we introduce a normalized Gray-Level Co-occurrence Matrix (GLCM) loss, in addition to the cross-entropy loss, thereby producing optimized results. This extends the problem into the following.
\begin{align}
    {<\theta_T, \theta_C, \theta_B, \theta_F>} = \argmin_{\theta_T, \theta_C, \theta_B, \theta_F} \Bigg( & \alpha \times \cel\Big(F(B(C(T(X, \theta_T), \theta_C), \theta_B), \theta_F), y\Big) + \notag\\ 
    & (1-\alpha) \times \glcml\Big(C(X, \theta_C), X\Big) \Bigg)
\end{align}
where $\alpha$ is a predefined weight. For GLCM loss, we first compute a GLCM matrix for the input before and after the colorization procedure, respectively. Then, the GLCM loss is expressed as the feature distance between the two matrices. Our objective is to minimize this distance since it represents the distortion introduced by colorization as compared to the original texture. This distance is formulated as a {\em L-infinite norm} which is equivalent to the equation below:
\begin{equation}
   \glcml =  \max_{1\leqslant i \leqslant m}\sum_{j=1}^{n}\Bigg|SOT\Bigg(grayscale\Big(C(X)\Big)\Bigg)_{i,j} - SOT(X)_{i,j}\Bigg|
\end{equation}
where SOT denotes {\em second-order texture} features of GLCM, $m$ is batch size and $n$ is the total number of SOT features. We consider the five SOT features described by \eqref{eq:asm}-\eqref{eq:dis} with a distance of 3 for 8 orientations (0/45/90/135/180/225/270 degrees). Hence,  $n=5\times 8 = 40$. All the 40 features are normalized in the same range when constructing the $m\times n$ SOT matrix.

\begin{figure*}[!ht]
\centering
\includegraphics[trim=7mm 3mm 1mm 4mm,clip,width=\textwidth]{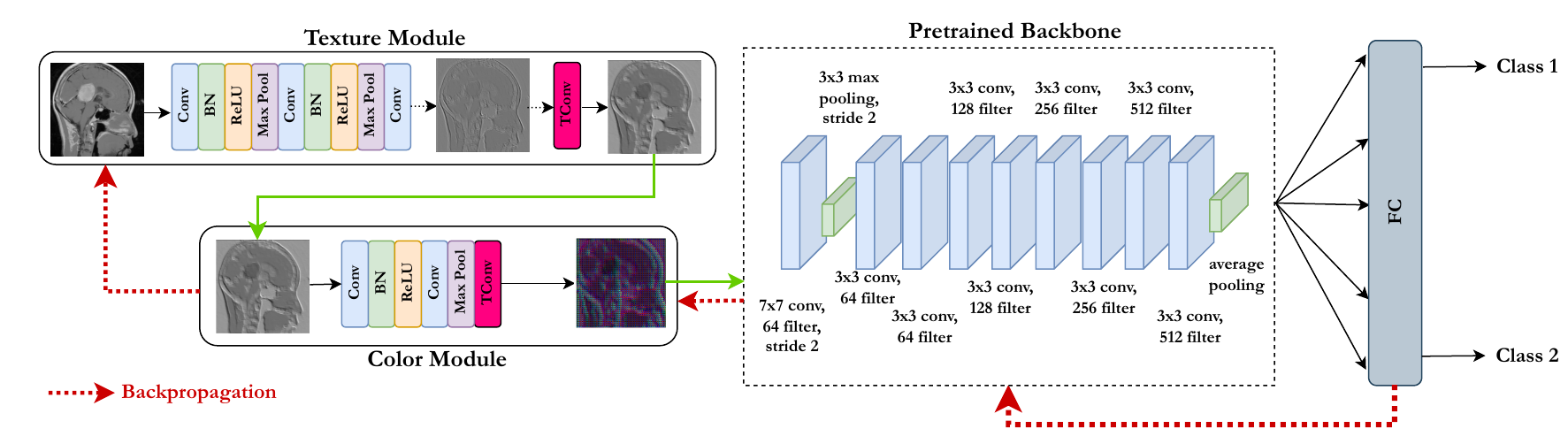}
\caption{Architecture of our proposed texture-color adaption alongside the backbone and final classifier.}
\label{fig:archi}
\end{figure*}

Our texture module follows a simplified autoencoder architecture, wherein the encoder comprises convolutional, batch normalization, ReLU, and max-pooling layers, ultimately generating a single-channel output. The decoder, on the other hand, utilizes transpose convolution operations to reconstruct the original input. The color module mirrors the structure of the texture module but adopts a shallower configuration. It outputs a three-channel image, subsequently fed into the pretrained backbone. For a visual representation of the architecture, refer to \fref{fig:archi}.

\section{Experiments}
\subsection{Dataset}
Our experiments encompassed four datasets representing distinct imaging modalities - Brain MRI\cite{1jny-g144-23}, Chest CT\cite{Alyasriy2021-vw}, Chest XRay\cite{Kermany2018-kd} and Breast Ultrasound\cite{ALDHABYANI2020104863}. Given the substantial class size imbalance inherent in medical images, we employed downsampling with a random sampling strategy to balance the datasets. Additionally, to address data scarcity concerns for specific classes and formulate a binary classification problem, we grouped various classes together. For example, within the Brain MRI dataset, we consolidate glioma, meningioma, and pituitary into a class designated as \textit{tumor}. See \tref{dataset} for an overview of the dataset.

\subsection{Setup}
Our dataset uniformity is maintained through the resizing and center-cropping of images, resulting in dimensions of (224, 224, 1), effectively reducing training costs. All experiments adopt a standardized hyperparameter set, including 300 epochs, a batch size of 32, a learning rate of 0.0001, and early stopping with a patience parameter set to 30. The experiments are conducted on GCP V100-SXM2-16GB with 4 GPUs to achieve a significant acceleration in processing. The experiment was conducted by performing the follow steps 1 and 2 independently, and then subjecting their output models to attacks as in step 3 for robustness comparison: {\bf 1) Fine-tune Base Models:} Three commonly-used pretrained models---ResNet18, ResNet50 \cite{he2015deep}, and DenseNet121 \cite{huang2018densely}---are adopted. Following \cite{Ma2021}, we replace the last layer of these three models by a new sequential layer consisting of a dense layer, a dropout layer, and a final dense layer. We then fine-tune these 3 models on our medical datasets.
{\bf 2) Train Base Models with our Texture and Color Modules:} The network architecture now consists of  of four components---Texture Module, Color Module, Pretrained Backbone, and Final Classifier---which are trained in an end-to-end fashion. GLCM and cross-entropy losses (with a predefined $\alpha$ = 0.98) are computed for each batch, and the overall loss is propagated backward through the entire network for updating parameters.
{\bf 3) Adversarial Attacks on Trained Models:} We launch various adversarial attacks on all the above models to compare performance degradation of basic transfer learning versus texture-color-adapted transfer learning, to assess the possible robustness improvement.
Note that both steps 1 and 2 use the same medical datasets, and all the models are fine-tuned/trained until nearly convergence, in order to ensure a fair comparison.

\subsection{Evaluation}
We assess the performance of various models based on their testing accuracy. The evaluation is conducted incrementally on a fine-tuned base model, the model with texture and color adaptation, and the model with texture and color adaptation combined with GLCM loss. To examine how different models respond to gradient-based adversarial attacks—specifically, FGSM, BIM, MIFGSM, and PGD—in terms of performance degradation, we subject our trained models to different attack perturbation sizes denoted as  $\epsilon$ (1/255, 2/255, 3/255, 4/255, 5/255, 6/255, 7/255, 8/255). The goal is to evaluate model performance under attacks with and without our proposed approach.

\begin{table}
\vspace{-5mm}
\centering
\caption{Dataset Overview}\label{dataset}
\begin{tabular}{|l|l|l|}
\hline
Dataset Name &  Classes & Class Size\\
\hline
Brain MRI &  {no-tumor, tumor (glioma, meningioma, pituitary)} & 1595, 1595\\
Chest XRay & {normal, pneumonia} & 1583, 1583\\
Chest CT &  {no-cancer (normal and benign), cancer} & 536, 536\\
Breast Ultrasound & {no-cancer (normal and benign), cancer} & 210, 210\\
\hline
\end{tabular}
\end{table}
\subsection{Results}
\begin{figure*}[!ht]
\centering
{\includegraphics[width=0.9\textwidth]{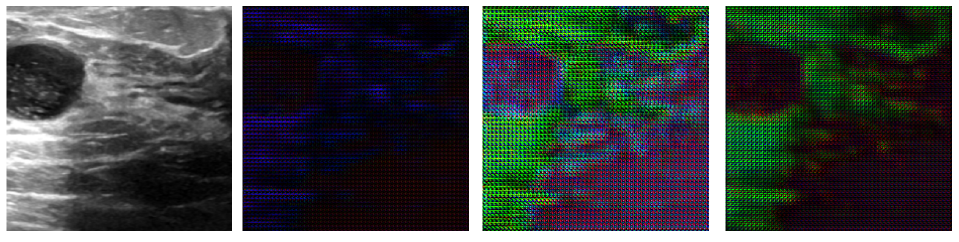}}%
\hfil
\caption{A demonstration of the incremental workflow. See from left to right: image for the original Breast Ultrasound image, sole colorization results, after adaptation of texture, result from addition of GLCM loss.}
\label{fig:usflow}
\end{figure*}

As depicted in Table \ref{acc_table}, our comprehensive experiments span four distinct medical imaging modalities—MRI, Ultrasound, CT, and X-Ray—employing three selected pretrained models: ResNet18, ResNet50, and DenseNet121. We compare the performance across different approaches. Notably, for all imaging modalities except Breast Ultrasound images, all three adapted approaches demonstrated comparable results compared to the fine-tuned base models. The incorporation of GLCM loss notably contribute to performance improvements or maintenance across most models. However, both the table and Figure \ref{fig:usflow} highlight that the adaptation of color and texture introduces distortion, and in some cases, undesirable noise to the original image. The limited size of the ultrasound dataset might as well contribute to the challenges in learning and the model's struggle to converge effectively. This effect was more pronounced in modalities such as Ultrasound, which contains more distinguishable and complex textures compared to other imaging modalities, resulting in a performance degradation. Our proposed GLCM loss has effectively mitigated this undesirable distortion, as evidenced by the results presented in \tref{acc_table}. Its efficacy is particularly evident on ultrasound compared to other imaging modalities, underscoring the criticality of preserving texture information.

\begin{figure*}[!ht]
\centering
\subfloat[Chest CT (ResNet18)]{\includegraphics[width=0.5\textwidth]{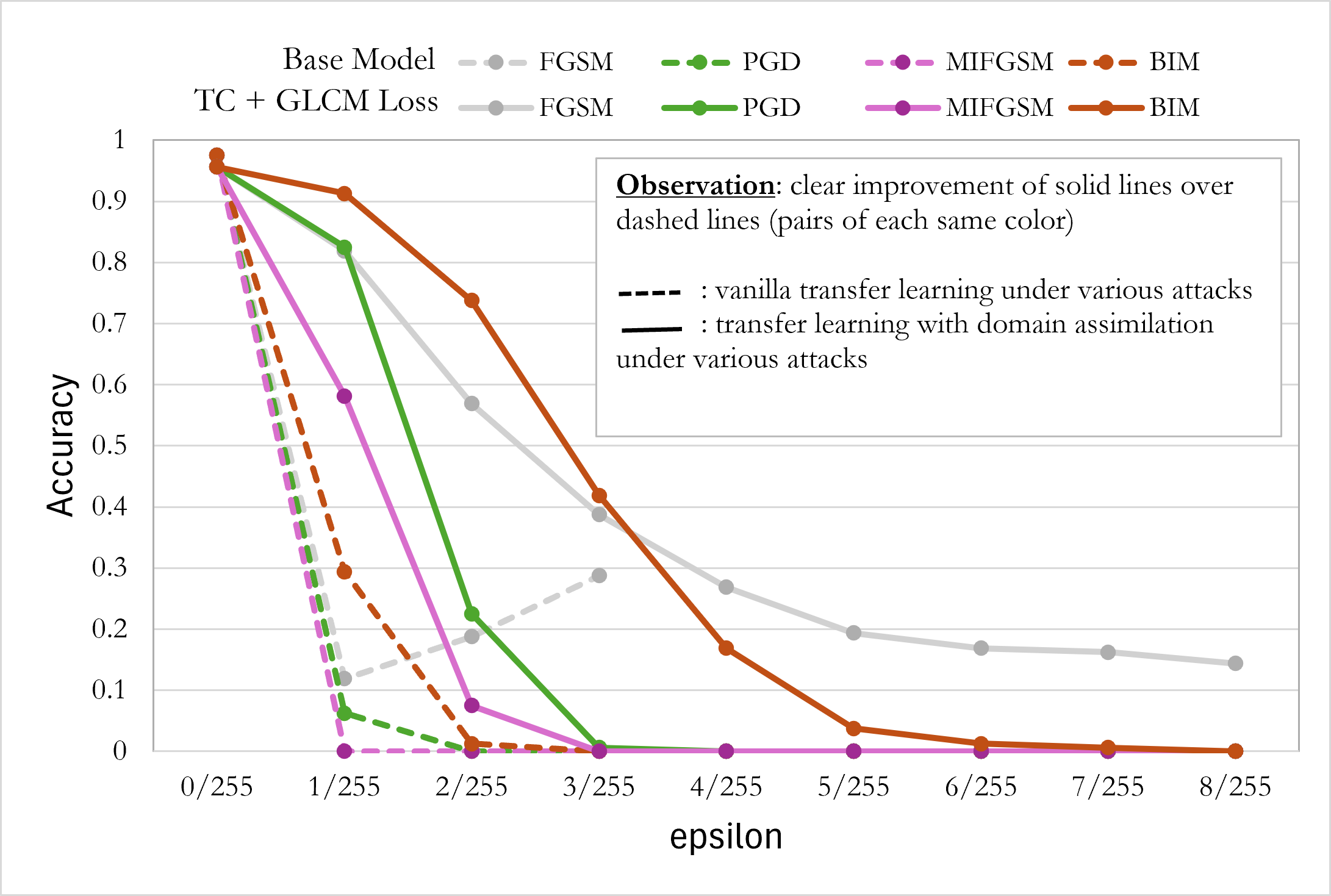}}%
\hfil
\subfloat[Breast Ultrasound (DenseNet121)]{\includegraphics[width=0.5\textwidth]{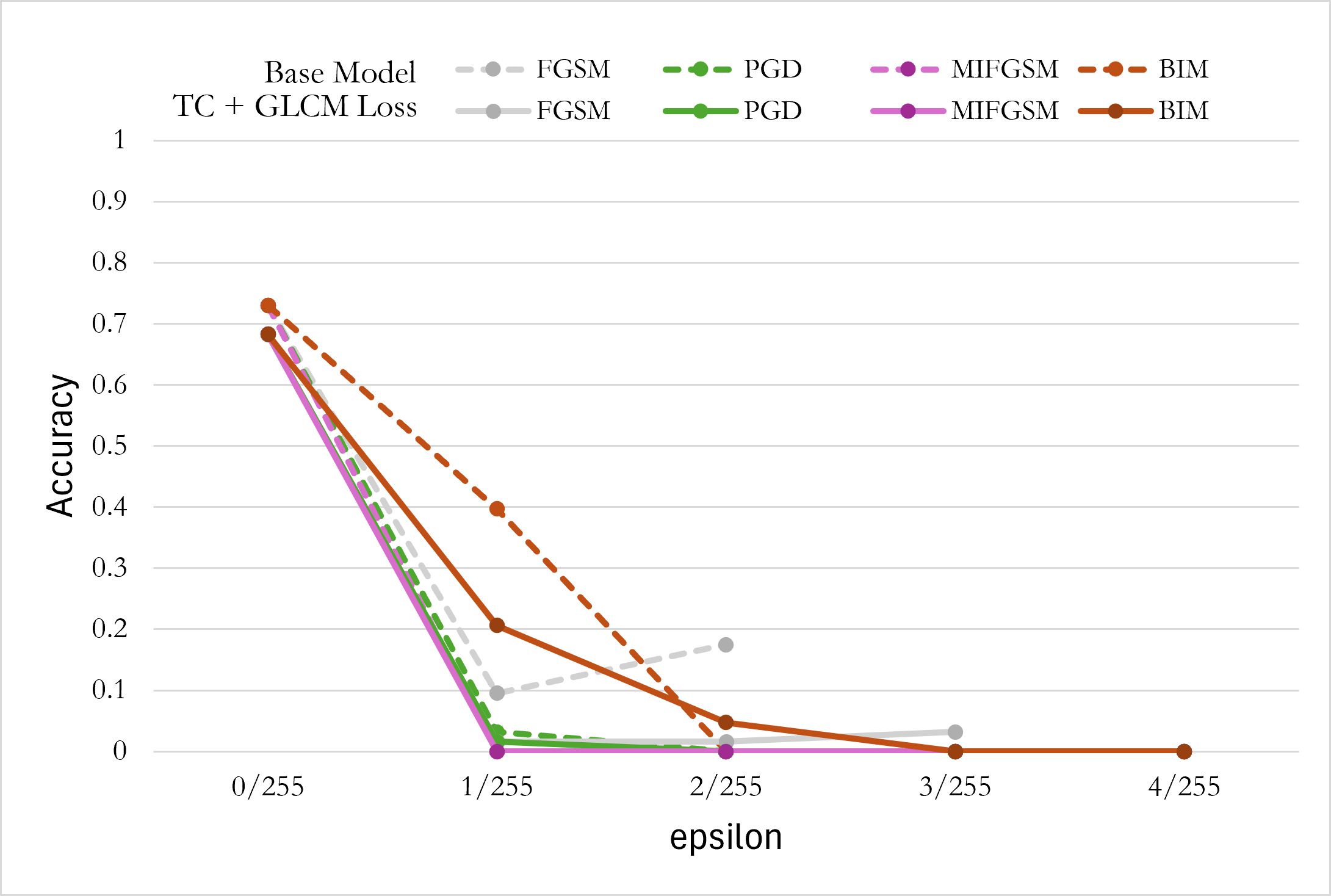}}%
\hfil
\caption{Robustness comparison under various adversarial attacks.}
\label{fig:aacomp}
\end{figure*}

Further in evaluating model robustness against gradient-based adversarial attacks—FGSM, BIM, MIFGSM, and PGD—we observe that our texture-color-adapted models with GLCM loss enhanced robustness by increasing the difficulty of generating successful attacks.  Figure \ref{fig:aacomp}a illustrates that the base model achieve <20\% accuracy under any attack (except for BIM) even with a very small perturbation magnitude (e.g., $\epsilon$ = 1/255); BIM reduces model accuracy to around 30\% at $\epsilon= 1/255$ but soon collapses the model with an accuracy of zero at $\epsilon= 2/255$. In contrast, armed with our domain assimilation strategy, the model's robustness increases notably: under all the attacks with $\epsilon=1/255$, the model maintains a reasonably good accuracy of about 90\%, 80\%, 80\%, and 60\% under the attacks BIM, PGD, FGSM, and MIFGSM, respectively. At a stronger attack strength of $\epsilon=3/255$, where the base model collapses to zero accuracy for almost all the attacks, our strategy helps maintain an accuracy of about 40\% under BIM and FGSM attacks. On the other hand, \fref{fig:aacomp}b shows that our approach was not effective for ultrasound. This can be attributed to the characteristic captured by Table \ref{acc_table}, which reveals that that ultrasound images contain more complex texture and hence were too sensitive to our adaptation. This calls for more advanced methods to be developed in future research.
The vanilla FGSM, represented by the dashed gray line in \ref{fig:aacomp}a and \ref{fig:aacomp}b, exhibits outlier behavior due to its characteristic of advancing gradients for a {\em single} step only. As such, augmenting the epsilon value would not enhance its efficacy but can potentially precipitate its significant deviation towards suboptimal solutions, as opposed to {\em iterative} gradient ascending as in other attack methods. Hence for FGSM, emphasis should be placed on very low epsilon values such as 1/255. 

\begin{figure*}[!ht]
\centering
{\includegraphics[width=0.7\textwidth]{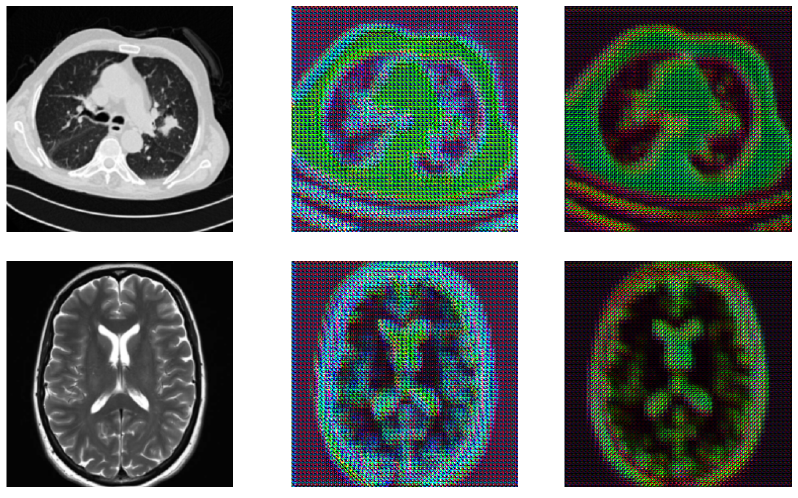}}%
\hfil
\caption{Transferred colorization from trained Breast Ultrasound color module to colorize given Chest CT and Brain MRI image. See first row for results for Chest CT and second row for Brain MRI. See from left to right: original image, colorization result from trained Breast Ultrasound color module, result with addition of GLCM loss.}
\label{fig:ccttransfer}
\end{figure*}

As part of our analysis, we also explored the transferability of learned parameters across different imaging modalities. Figure \ref{fig:ccttransfer} illustrates the transferred colorization from a model trained on Breast Ultrasound images to colorize Chest CT and Brain MRI images. The results suggest that our adapted models can be effectively transferred to different modalities. With further fine-tuning, we believe these models can be even better adapted to diverse datasets. 

\begin{table*}[!ht]
\centering
\caption{Model Accuracy before \& after Adaptation w/ and w/o Correction (by GLCM loss). No Attack.}\label{perf}
\begin{tabular}{|ccc|c|c|c|c|}
\hline
\multicolumn{3}{|c|}{} & \textbf{Brain MRI} & \textbf{Breast {Ultrasound}} & \textbf{Chest CT} & \textbf{Chest X-Ray} \\ \hline
\multicolumn{1}{|c|}{\multirow{4}{*}{\textbf{ResNet18}}}    & \multicolumn{2}{c|}{Base w/ Fine Tuning}                      & 97.9\%            & 74.6\%            & 97.5\%           & 92.2\%             \\ \cline{2-7} 
\multicolumn{1}{|c|}{}                                      & \multicolumn{1}{c|}{\multirow{2}{*}{Adapted}}                       & Texture-Color (TC) & {96.4\%}            & 71.4\%            & {95.6\%}           & {90.9\%}             \\ \cline{3-7} 
\multicolumn{1}{|c|}{}                                      & \multicolumn{1}{c|}{}                         & TC+GLCMLoss   & {96.9\%}            & 71.4\%            &{95.6\%}                   &{91.8\%}                     \\ \hline
\multicolumn{1}{|c|}{\multirow{4}{*}{\textbf{ResNet50}}}    & \multicolumn{2}{c|}{Base w/ Fine Tuning}                      & 97.5\%            & 79.4\%            & 97.5\%           & 92.0\%             \\ \cline{2-7} 
\multicolumn{1}{|c|}{}                                      & \multicolumn{1}{c|}{\multirow{2}{*}{Adapted}}                          & Texture-Color (TC) & {97.3\%}            & 76.2\%            & {96.3\%}           & {91.6\%}             \\ \cline{3-7} 
\multicolumn{1}{|c|}{}                                      & \multicolumn{1}{c|}{}                         & TC+GLCMLoss   &{97.3\%}                    & 77.8\%            &{96.9\%}                   & {91.8\%}                     \\ \hline
\multicolumn{1}{|c|}{\multirow{4}{*}{\textbf{DenseNet121}}} & \multicolumn{2}{c|}{Base w/ Fine Tuning}                      & 98.1\%            & 73.0\%            & 98.8\%           & 92.6\%             \\ \cline{2-7} 
\multicolumn{1}{|c|}{}                                      & \multicolumn{1}{c|}{\multirow{3}{*}{Adapted}}                         & Texture-Color (TC) & 97.5\%            & 66.7\%            & {98.1\%}           & {92.2\%}                     \\ \cline{3-7} 
\multicolumn{1}{|c|}{}                                      & \multicolumn{1}{c|}{}                         & TC+GLCMLoss   &97.3\%                    & 68.3\%            &{98.1\%}                   & {92.2\%}                    \\ \hline
\end{tabular}
\label{acc_table}
\end{table*}

\section{Conclusion}
In this study, we introduce texture and color adaption into transfer learning to bridge the domain discrepancy existing in AI-based medical imaging, consequently addressing the heightened vulnerability of medical AI models to adversarial attacks. Our proposed approach consists of a texture-color adaptation module that dynamically learns parameters in conjunction with pre-trained models, and a GLCM loss that retains essential texture information to restrict distortion, fostering a resilient model over various imaging modalities. Our evaluation demonstrate enhanced model robustness against adversarial attacks, specifically gradient-based adversarial examples created by BIM, PGD, FGSM, and MIFGSM. Our analysis also discover the challenges posed by imaging modalities with intricate textures, exemplified by Ultrasound images. Our findings validate the domain assimilation idea and the effectiveness of the tamed adaption approach, yet also pointing out potential future work on improvement for different imaging modalities. 


\bibliographystyle{splncs04}
\bibliography{aami}

\end{document}